%% file: main.tex
\documentclass[10pt,journal,compsoc]{IEEEtran}

\title{Filler}
\author{Filler}

\ifCLASSOPTIONcompsoc
  \usepackage[nocompress, noadjust]{cite}
\else
  \usepackage{cite}
\fi

\ifCLASSINFOpdf
  \usepackage[pdftex]{graphicx}
\else
   \usepackage[dvips]{graphicx}
\fi
\usepackage{amsmath, bm}
\usepackage{acronym}
\usepackage{array}

\usepackage{mdwmath}
\usepackage{mdwtab}

\usepackage{eqparbox}

\ifCLASSOPTIONcompsoc
  \usepackage[caption=false,font=footnotesize,labelfont=sf,textfont=sf]{subfig}
\else
  \usepackage[caption=false,font=footnotesize]{subfig}
\fi

\usepackage{stfloats}
\usepackage{url}
\newcommand\MYhyperrefoptions{bookmarks=true,bookmarksnumbered=true,
pdfpagemode={UseOutlines},plainpages=false,pdfpagelabels=true,
colorlinks=true,linkcolor={black},citecolor={black},urlcolor={black},
pdftitle={Bare Demo of IEEEtran.cls for Computer Society Journals},
pdfsubject={Typesetting},
pdfauthor={Michael D. Shell},
pdfkeywords={Computer Society, IEEEtran, journal, LaTeX, paper,
             template}}
\ifCLASSINFOpdf
\usepackage[\MYhyperrefoptions,pdftex]{hyperref}
\else
\usepackage[\MYhyperrefoptions,breaklinks=true,dvips]{hyperref}
\usepackage{breakurl}
\fi
\usepackage{lineno,hyperref}
\usepackage{placeins}
\usepackage{graphicx}
\usepackage{booktabs}
\usepackage{multirow}
\usepackage{siunitx}
\usepackage{adjustbox}

\usepackage{tikz,xcolor}

\definecolor{lime}{HTML}{A6CE39}
\DeclareRobustCommand{\orcidicon}{
	\begin{tikzpicture}
	\draw[lime, fill=lime] (0,0) 
	circle [radius=0.16] 
	node[white] {{\fontfamily{qag}\selectfont \tiny ID}};
	\draw[white, fill=white] (-0.0625,0.095) 
	circle [radius=0.007];
	\end{tikzpicture}
	\hspace{-2mm}
}

\foreach \x in {A, ..., Z}{\expandafter\xdef\csname orcid\x\endcsname{\noexpand\href{https://orcid.org/\csname orcidauthor\x\endcsname}
			{\noexpand\orcidicon}}
}

\newcommand{\mathdash}{\mathit{\textnormal{-}}}
\newcommand{\mse}{\mathit{MSE}}
\newcommand{\rmse}{\mathit{RMSE}}
\newcommand{\f}{\mathit{F_1}\textnormal{-}\mathit{score}}

\begin{document}
%
\title{Road Roughness Estimation Using Machine Learning}
%
%
%
%

\author{Milena~Bajic\orcidM{},
        Shahrzad~M.~Pour\orcidS{},
        Asmus~Skar\orcidA{},
        Matteo~Pettinari\orcidP{},
        Eyal~Levenberg\orcidE{},
        Tommy~Sonne~Alstrøm\orcidT{}
    \IEEEcompsocitemizethanks{\IEEEcompsocthanksitem M. Bajic, T.S. Alstrøm and S.M. Pour are with the Department of Applied Mathematics and Computer Science, Technical University of Denmark (DTU), Denmark.\protect%
    \IEEEcompsocthanksitem A. Skar and E. Levenberg are with the Department of Civil Engineering, Technical University of Denmark (DTU), Denmark.\protect%
    \IEEEcompsocthanksitem M. Pettinari is with the Vejdirektoratet, Denmark.\protect
    \IEEEcompsocthanksitem Corresponding author: Milena Bajic (E-mail: mibaj@dtu.dk)}
    \thanks{This work has been submitted to the IEEE for possible publication. Copyright may be transferred without notice, after which this version may no longer be accessible.}
    }

%
%

\markboth{}
{Bajic \MakeLowercase{\textit{et al.}}: Road Roughness Estimation Using Machine Learning}
%



\IEEEtitleabstractindextext{%
\begin{abstract}
Road roughness is a very important road condition for the infrastructure, as the roughness affects both the safety and ride comfort of passengers. The roads deteriorate over time which means the road roughness must be continuously monitored in order to have an accurate understand of the condition of the road infrastructure. In this paper, we propose a machine learning pipeline for road roughness prediction using the vertical acceleration of the car and the car speed. We compared well-known supervised machine learning models such as linear regression, naive Bayes, k-nearest neighbor, random forest, support vector machine, and the multi-layer perceptron neural network. The models are trained on an optimally selected set of features computed in the temporal and statistical domain. The results demonstrate that machine learning methods can accurately predict road roughness, using the recordings of the cost approachable in-vehicle sensors installed in conventional passenger cars. Our findings demonstrate that the technology is well suited to meet future pavement condition monitoring, by enabling continuous monitoring of a wide road network.

\end{abstract}

}

\maketitle

\IEEEdisplaynontitleabstractindextext

%
\IEEEpeerreviewmaketitle

\input{chapters/introduction}
\input{chapters/methods}

\input{chapters/results}
\input{chapters/conclusion}
\fnbelowfloat



%



\ifCLASSOPTIONcompsoc
  \section*{Acknowledgments}
\else
  \section*{Acknowledgment}
\fi

The authors would like to thank the Innovation Fund Denmark \cite{ifd}, who supported this work with the grant ``Live Road Assessment tool based on modern car sensors LiRA'' .

\ifCLASSOPTIONcaptionsoff
  \newpage
\fi




\bibliographystyle{IEEEtran}
\bibliography{bibliography}








\end{document}

%% file: chapters/introduction.tex
\ifCLASSOPTIONcompsoc
\IEEEraisesectionheading{\section{Introduction}\label{sec:introduction}}
\else
\section{Introduction}
\label{sec:introduction}
\fi



\IEEEPARstart{T}{he} road infrastructure is a basic physical asset essential for the facilitation of society and for generating economic growth and development \cite{sharif2021effects}. With time and exposure to traffic and climate loading, the road conditions continuously change and deteriorate affecting the safety of drivers and passengers, ride comfort, handling stability, energy consumption and vehicle maintenance costs \cite{radovic2016measurement}. Thus, maintenance of the road infrastructure is one of the primary concerns of road administrations. Maintenance is typically performed in a reactive (i.e. repairing damaged road sections) or preventive manner (i.e. repairing road sections that are expected to deteriorate rapidly). Both strategies require regular monitoring of the road network condition, identifying imminent or developing damage.

In order to support decision-makers in optimizing repair and maintenance efforts several road condition indicators have been developed \cite{shah2013a}. In this regard, the road roughness is considered most important, as it does not only affect the pavement’s ride quality, but also energy consumption, vehicle delays and maintenance costs \cite{janani2020a}. The International Roughness Index (IRI) is the most common index applied worldwide for characterizing longitudinal road roughness and managing road systems. IRI serves as a measure of pavement performance and ride quality, and is highly correlated with the overall ride- and pavement loading vibration level \cite{sayers1998a}.

The IRI is calculated using the algorithm proposed by \cite{sayers1986guidelines}, which essentially provides the response of a mathematical quarter-car vehicle model moving over a road profile. IRI is based on the simulation of the roughness response of a car travelling at 80 km/h - it is the reference average rectified slope, which is a ratio of a vehicle's accumulated vertical suspension motion divided by the distance traveled. A detailed overview of IRI specifications worldwide can be found in \cite{mucka2017a}. 

Calculation of IRI requires measurements of the longitudinal road profile, typically in each wheel-track. Standard methods for this purpose include manual inspection surveys (e.g., rod and level and walking profilometers) and automated inspection surveys involving specially equipped vehicles (i.e., inertial sensors). These measurements provide accurate and detailed spatial measurements but often lack the required surveying frequency. Thus, defects that progress faster than the current surveying frequency will escape detection. 

In recent years, several new concepts have been proposed to overcome some of the limitations of standard road condition monitoring methods, utilizing vehicles as sensing platforms. The approach for doing so have mainly focused on crowdsensing methods, using inertial sensors in smartphones \cite{seraj2015roads, alessandroni2017a, sattar2018road} or custom-made sensor platforms \cite{eriksson2008a,masino2017a,el-wakeel2018a}. In this paper, both built-in vehicle sensors and external embedded sensors are utilized, a concept recently proposed in \cite{levenberg2021lira}. Some of the advantages with this approach over others include: (i) vehicle information is known, (ii) sensors and vehicle is well coupled and (iii) vehicle information and sensor data is synchronized.

The relation between vehicle vibrations and the road profile have traditionally been investigated utilizing dynamic response models  \cite{imine2006a,gonzalez2008a, doumiati2011a,fauriat2016a}. However, the difficulties associated with the practical implementation of such a methodology may include: (i) determination of system parameters, (ii) model formulation, (iii) model inversion, and (iv) sensitivity to changes in system parameters with time.

Another approach, which is the focus of this paper, is road roughness prediction utilizing machine learning (ML) methods which have the advantage that vehicle system parameters are not needed a priori.

A Bayesian-regularized NARX neural network for road profile prediction from simulated vehicle response, under varying vehicle speeds, payloads and noise was proposed in \cite{ngwangwa2010reconstruction}. The network was trained and evaluated on simulated artificially generated road profiles. The study showed possibility of road roughness prediction, using the well controlled data, under known conditions.

A multi-layer perceptron (MLP) artificial neural network, trained using the simulated response of a full vehicle model of the Land Rover, at measured road profiles with minor corrections was developed in \cite{ngwangwa2014reconstruction}. The model was tested on data driven over discrete obstacles and Belgian paving, resulting in $\rmse$ of \si{15}{\%}. \cite{islam2014use,forslof2015roadroid,du2016application} developed linear regression models using acceleration to predict IRI and \cite{douangphachanh2014exploring} showed that the addition of gyroscope and speed data into the linear model improves the prediction. \cite{belzowski2015evaluating} developed linear and logistic regression models to predict the IRI value and category and demonstrated higher accuracy when predicting the IRI categories. 

A comparison of different ML models for the weighted longitudinal profile prediction, using simulated dynamic response of a vehicle traversing over a measured road profile was done in \cite{nitsche2014comparison}. They compared the performance of the multilayer perceptron (MLP), support vector regression (SVR) and random forests (RF), with the SVR resulting in the best performance, although the models achieved a similar performance. As inputs, they computed features in the statistical and the frequency domain from the 3D acceleration and the wheel's angular velocity. The models were tested on data recorded with a probe vehicle and a target measured using an inertial profilometer. The results showed the effectiveness of machine learning for the weighted longitudinal profile prediction and while the models were trained using the simulated data, they were validated on measured data, however without including the effect of varying speed. 

A study of IRI prediction using the simulated signals of connected vehicles traversing over the same road segments was done in \cite{zhang2018application}. They applied a convolution operation to extract common road related features, while suppressing the impact of different mechanical properties and speeds of vehicles. The extracted features were used as input features to an ANN model which resulted in performances with $\f$ range of $0.94-0.98$ for different IRI levels. 

A methodology for IRI prediction, based on a half-car model simulation of vehicle dynamic response, at measured road profile and realistic driving speed profiles was proposed in \cite{jeong2020convolutional}. Simulated are the responses of four vehicle types, under five different driving speed profiles and explored were multiple combinations of sensors, namely the vertical acceleration, driving speed and angular velocity. They represented inputs as 2D images and utilised a 2D-convolutional neural network (CNN) model, typically applied on image data. They explored the CNN performance under different combinations of vehicles, inputs and driving speeds and concluded that they showed similar performances with the $\rmse$ errors of $0.5-0.6$ \SI{}{\meter\per\kilo\meter}

A detailed study on variables affecting the road roughness prediction using a smartphone inside of a moving vehicle, in state-of-the-art simulation settings was conveyed in \cite{wang2020study}. They studied the impact of different vehicle types and their mechanical properties and the driving styles on dynamical responses, as well as the position and the smartphone type. They devoleped multiple machine learning models for IRI, obtaining the best results of $\rmse$ between 0.73 between 0.91 \SI{}{\meter\per\kilo\meter} using the MLP model.

A random forest regression model for IRI prediction from the initial IRI (IRI at the moment of pavement construction) along with the rutting depth, road distress, traffic, climate and pavement structure variables, was proposed in \cite{gong2018use} and obtained $R^{2}=0.974$ on test data. \cite{abdelaziz2020international} proposed MLP model using the initial IRI, pavement age, rutting depth and cracks data and obtained $R^{2} = 0.75$. A study using the MLP and climate and traffic parameters as inputs was conveyed in \cite{hossain2020artificial}. A study of the impact of different SVM kernels for IRI regression task using pavement structure, climate changes, and traffic loadings as input variables was performed in \cite{ziari2016prediction} and the best performance was obtained using the Pearson III Universal kernel.

While all these methods show merit, they are typically based on smartphone measurements, without a wider and calibrated usage, or developed in simulated settings and hence dependent on the complexity of the simulation. The purpose of this study is to explore the usage of machine learning methods for IRI prediction, in a realistic manner, using the data recorded by cars, at varying speeds, hence tacking the inevitable problems occurring in such setup. Hence, this study aims at aiding the transition efforts from traditional road roughness condition surveys, characterized by information gaps of the order of years, to real-time monitoring of wide road network utilizing in-vehicle sensors. 

The proposed pipeline and the analysis method for performing road roughness prediction from the collected sensor readings is described in Section \ref{sec_method}, followed by the results and conclusions presented in Sections \ref{sec_results} and \ref{sec_conc}, respectively. Overall, the obtained results provide confidence in the proposed methodology, and demonstrate that the technology is suited to meet future pavement condition evaluation needs.

%% file: chapters/methods.tex
\section{Methodology}
\label{sec_method}

\subsection{Data Collection}
The data collection was carried out using Green Mobility (GM) \cite{GM} Renault Zoe electric cars, equipped with an IoT hardware unit - AutoPi Telematics Unit with an embedded Raspberry Pi unit including the GPS and accelerometer modules as described in \cite{levenberg2021lira}.

In this study, we utilize geo-referenced location data collected using the GPS sensor sampled at \SI{1}{\hertz} with \SI{3}{\meter} accuracy, vertical vibrations data via accelerometer sampled at \SI{50}{\hertz} with a \SI{3}{\milli\gram} accuracy and speed collected using the built-in vehicle sensor. The data is collected on \SI{52.6}{\km} route, covering the both lanes of M3 Highway in Copenhagen.

The GM data is labeled using the reference data, recorded by the the Greenwood Profiler and operated by the Danish Road Directorate, providing the IRI measurement per every \SI{10}{\metre} segment. 

\subsection{Proposed Framework}
\begin{figure*}[!t]
\centering
\includegraphics[width=\linewidth]{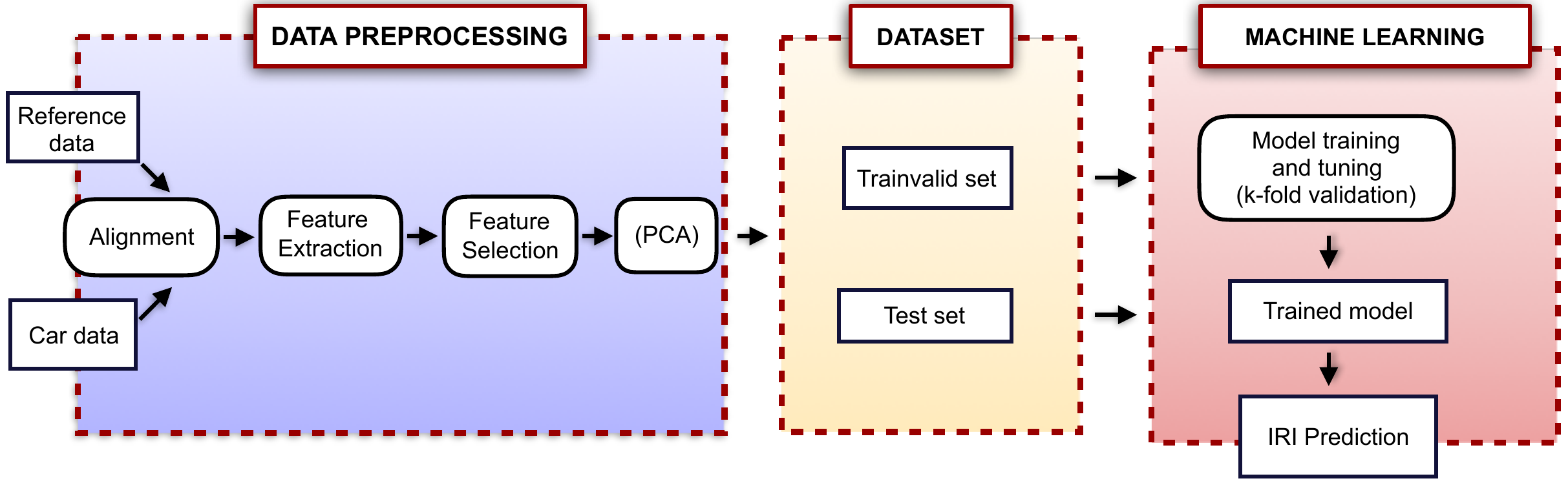}
\caption{Overview of the proposed framework.}
\label{fig:fw}        
\end{figure*}

The proposed framework for estimating the IRI is based on the analysis of the $(a_{z}, v)$, namely the acceleration in the z-direction and the car speed.  The framework consists of three phases: i) data preprocessing, ii) data preparation and iii) machine learning, as illustrated in Fig. \ref{fig:fw}. 

In the preprocessing phase, the car data is geographically aligned with the reference data, providing the IRI labels for machine learning and an optimal subset of discriminative features is computed. Subsequently, the prepared dataset is split into a set for training and validation of the machine learning models and a holdout test dataset for performance evaluation. In the last phase, a set of supervised machine learning models is trained to predict the IRI label from the optimal feature set and the performance is tested on the test dataset. 

\subsection{Segment Alignment}
The alignment between the GM and the reference data is performed in three stages: i) map matching where the GPS recorded routes are corrected, using the road network information, ii) interpolation where GPS coordinates are assigned to all sensors and iii) segment alignment where the closest matching GM and referenced data segments are matched, resulting in labeled GM segments.

In the first stage, the vehicle routes, obtained using the GPS sensors are map matched to the OSM road network \cite{haklay2008openstreetmap}, using the Open Source Routing Machine (OSRM) \cite{luxen-vetter-2011}. The OSM is a collaborative map project which freely shares the geodata. The OSRM is a web-based network service that uses the road data from OSM and has implemented several services mainly for routing purposes but also the map matching service. The map matching is done using the Hidden Markov model (HMM). According to the HMM applied on the map matching problem \cite{newson2009hidden}, the hidden states are defined as road segments, while the observed states are defined as GPS observations. Finally, the HMM is employed to detect the most probable sequence of road segments while taking into account the GPS measurement noise and the road network.

In the next stage, the GPS geolocations are associated to all sensor measurements by performing interpolation between the adjacent GPS measurements, assuming the constant vehicle speed between them.

In the final stage, the starting and the ending point of each reference segment are matched to the closest interpolated car points and the measurements in between are assigned to a matching segment. Additionally, the data is cleaned by applying the requirement of $\ge2$ points per segment. The alignment error is reduced by averaging the IRI over \SI{100}{\metre} and the number of segments is increased, by applying a sliding \SI{100}{\metre} window with a step of \SI{10}{\metre} is applied. 

\subsection{Feature Extraction}
In the feature extraction stage, a set of 68 features is computed in the statistical and the temporal domain per segment and the optimal subset of relevant features is further described in Section \ref{sec_results}. Initially, the segment lengths are equalized by resampling them using the linear interpolation between the closest points.

The features are computed using the tsfel library \cite{barandas2020tsfel} and subsequently standardized, namely centered at the mean value of $0$ and scaled to the standard deviation of $1$.

\subsection{Feature Selection}
Irrelevant features, namely the features uncorrelated with the target and redundant features which carry the information already present in the subset, increase the noise in the data, making the training process harder. Hence, removal or irrelevant and redundant features is an important step in machine learning. 

In our analysis, feature selection is employed in two stages. In the first stage, the constant features are discarded as they do not provide information that is useful for discrimination between different cases. In the second stage, the optimal subset of irredundant input features is selected using the forward Sequential Feature Selection (SFS) algorithm \cite{aha1996comparative}. 

The SFS belongs to the class of wrapper methods which evaluate possible feature subsets, while taking feature interactions into account. The SFS selects the subset which results in the best performance in terms of the chosen metrics, evaluated using a specific ML model. The SFS starts by evaluating all features individually and selects the feature resulting in the best performance. In the following stages, the combinations with the remaining variables are tested and at each step, the variable which results in the largest performance improvement is selected. Hence, the performance improvement is related to the subset, meaning that if a feature is independently informative, it will not be selected if the provided information is already present. 

The SFS performance is evaluated using the random forest model and optimized to minimize the $RMSE$, in a 5-fold cross-validation manner. The algorithm is implemented the mlxtend library \cite{raschkas_2018_mlxtend}.

\subsection{Principal Component Analysis}
While all the features in the subset obtained using the SFS provide new information, the correlations between them are still possible. Hence, we additionally test our models on a decorrelated subset obtained using the Principal Component Analysis (PCA) \cite{massy1965principal} applied after the SFS. 

The PCA is a dimensional reduction technique where the initial feature set is transformed into a reduced set of features, e.g. principal components. An advantage of the PCA transformed feature set is that a smaller number of uncorrelated principal components can be selected to explain a large percent of the data variance. In our analysis, the number of PCA components is selected to explain $99\%$ of the variance. 

\subsection{Machine Learning Algorithms}
The objective of supervised ML models if to find a function $f(x;\boldsymbol{\theta})$ which predicts the real-valued or categorical target $\bm{y} = (y_1,..,y_N)$ from a set $X = (\bm{x}_1,..,\bm{x}_N)$:
\begin{equation}
\label{eqn_sl}
\bm{y} = f(X;\boldsymbol{\theta})
\end{equation}
where $\bm{x_i}$ denotes the feature vector of for sample $i$, $N$ the number of samples and $\bm{\theta}$ are the model parameters. The $\bm{\theta}$ vector is estimated to minimize the cost function, defined to measure the discrepancy between the predicted  $\bm{\hat{y}} = (\hat{y}_1,..,\hat{y}_N)$ and the actual values $\bm{y} = (y_1,..,y_N)$. 

In regression, we use the the mean squared loss $\mse$ as the cost function:
\begin{equation}
\label{eqn_mse}
\mse = \frac{1}{N}\sum_i^N (y_i - \hat{y}_i)^2
\end{equation}

In classification, we use the categorical cross entropy $CE$ as the cost function:
\begin{equation}
\label{eqn_ce}
CE = -\frac{1}{N}\sum_{i}^{N}\sum_{j}^{C}  y_{ij} \mathrm{log}{\hat{y}_{ij}}
\end{equation}
where $C$ denotes the number of classes of the categorical target $\bm{y}$. The cost functions are optimized in the training stage, but if the model has too high flexibility, it can begin to adapt to the noise in the training data instead of the general patterns. This leads to overfitting, which negatively impacts the generalization error. To handle this problem, additional regularization terms can be introduced into the cost function, which act as a penalty on model parameters and favors smoother models. Commonly applied regularization terms are the $L_1$ and $L_2$ norm on the model weights which penalize the sum of absolute and squared model weights respectively: 
\begin{align}
\label{eqn_l1l2}
L_1 = \lambda  \sum_{j=1}^n |\theta_j|\\
L_2 = \lambda  \sum_{j=1}^n \theta^2_j 
\end{align}
where $n$ denotes the number of model parameters. 

The model parameters which are not directly optimized in training procedure are denoted as hyperparameters. In our analysis, the model hyperparameters are tuned using the grid search with 5-fold cross-validation. Standard cross-validation procedure is employed with random split, but due to the correlations between the adjacent road segments, the observations in our analysis are not independent and random split is not appropriate. It has been observed that this procedure leads to overfitting. Hence, the data is split into folds while preserving the ordering of segments. 

In the classification task, we developed models for the prediction of the IRI level determined by the Danish Road Directorate as described in Table \ref{table:iri_classes}. The higher IRI levels refer to worse road states and are less frequent in our dataset. The data is resampled by generating synthetic samples for minority classes using the Adasyn algorithm \cite{he2008adasyn} in which the class examples that are harder to learn are generated more frequently compared to the examples easier to learn. The Adasyn algorithm is implemented using imblearn \cite{JMLR:v18:16-365}. 

\begin{table}[!t]
\renewcommand{\arraystretch}{1.3}
\caption{IRI Level Definition.}
\label{table:iri_classes}
\centering
\begin{tabular}{llll}
\toprule
 & IRI $\leq$ 0.9 &  0.9 $<$ IRI $\leq$ 1.6 & IRI $>$ 2.5 \\
\midrule
\textbf{IRI level} & Low        & Medium       &  High \\
\bottomrule
\end{tabular}
\end{table}

The ML performance is compared to a simple baseline model, defined by always predicting the mean IRI (regression) or the most frequent class (classification) of the training dataset. In our analysis, considered are the i) linear and regularized linear models, ii) k-Nearest Neighbors, iii) Random Forest, iv) Support Vector Machine and v) Multi-Layer Perceptron Neural Networks. In classification, additionally considered is the Naive Bayes model. The models are implemented using the sklearn library \cite{sklearn_api}. 

\subsubsection{Linear Models}
Employed are the multiple linear and regularized linear models, namely Ridge, Lasso and elastic net in regression and logistic regression in classification modelling. 

The simplest model is a multiple linear model where the target $\bm{y}$ is modelled as a linear combination of input features:
\begin{equation}
\label{eqn_lin}
\bm{y} = \bm{\theta} X
\end{equation}

An advantage of the linear model is its simplicity and interpretability. It performs well when there is a linear relationship between the target and the input features and in the absence of multicollinearity and extreme outliers. However, when these conditions are not met, it leads to unstable solution and poor performance. 

As mentioned earlier, to combat the overfitting problem, commonly employed are the regularized linear models, such as Lasso, Ridge and the elastic net, constructed by the linear combination of the L1 and L2 regularization terms \cite{de2009elastic}.


In classification, employed is the logistic regression model, which maps the continuous output of the linear model to a class probability, by applying the logit function in binary and categorical cross-entropy or logit with the one-vs-rest scheme in multiclass classification. In our analysis, logistic regression model is implemented with the L2 regularization and the L2 regularization strength is tuned in grid search.

\subsubsection{k-Nearest Neighbors}
The k-nearest neighbors (kNN) \cite{fix1951nonparametric, altman1992introduction} is a simple, nonparametric model where the prediction is based on the surrounding neighborhood in the training set. The prediction is computed as the mean value (regression) or the most frequent class (classification) of the $k$ nearest data points. The $k$ hyperparameter is an important hyperparameter, significantly affecting the model performance. Small $k$ values can cause overfitting, leading to a large generalization error, while high values can cause underfitting and increase the bias error. 

\subsubsection{Naive Bayes}
Naive Bayes is a simple probabilistic classification model based on Bayes theorem. The probability of a class $c_k$, given the observed values of input features $P(c_k|\bm{x})$ is computed as:
\begin{equation}
\label{eqn_bayes}
P(c_k|\bm{x}) = \frac{ P(\bm{x}|c_k) P(c_k)}{P(\bm{x})}
\end{equation}
where $\bm{x}$ is the feature vector, $P(\bm{x}|c_k)$ is the probability to obtain $\bm{x}$ given class $c_k$, $P(c_k)$ is the probability to observe class $c_k$ and $P(\bm{x})$ is the evidence. In \eqref{eqn_bayes}, there is an "naive" assumption of independence between input features.

\subsubsection{Random Forest}
Random forest (RF) \cite{breiman2001random} is a powerful model formed by an ensemble of decision trees, each trained on a random subsample using a random subset of input features, which reduces overfitting and increases the performance. Random forests can be employed as a classification model, where the predicted class is decided by majority voting or as a regression model where the predicted target is computed as the mean of all predicted targets of the individual trees. 

The trees are trained by making selections on input features which results in partitions of the data into smaller and more homogeneous groups. A tree consists of multiple nodes where at each node, a split is made based on the value of one specific feature which leads to the highest impurity decrease. In our analysis, the impurity decrease is defined as the decrease in the $\mse$ for regression, and for classification we use the $\mathit{Gini}$ index, defined as:
\begin{equation}
\label{eqn_gini}
Gini = 1-\sum_{i=1}^C p^2_i
\end{equation}
where $C$ is the number of classes and $p_i$ is the probability of class $i$. 

The trees are grown on using random samples with replacement, resulting in different training subsamples for different trees.

A number of hyperparameters is tuned in grid search, namely the number of the trees, the number of input features considered at every split and the maximum depth of the trees.

\subsubsection{Support Vector Machine}
Support vector machine (SVM) is a powerful model, initially proposed for classification and later developed for regression problems \cite{vapnik1997support}. The objective of the SVM is to maximize the margin of the decision boundary between different classes. Since many real cases are not fully separable, violation of the margin is permitted but penalized by a hyperparameter $C$. Smaller values of $C$ allows for more margin violations resulting in simpler models while larger values allow less violations and are more likely to overfit the data. Hence, the tuning of the parameter $C$ directly impacts the bias-variance tradeoff.

For regression, a parameter, $\epsilon$, is introduced which characterizes the margin of tolerance. Predictions with a deviation larger than the $\epsilon$ are penalized ($\epsilon$-insensitive function). Hence, the objective of SVM for regression task is to minimize the $\epsilon$-insensitive function $\mathcal{L}_{\epsilon}$:
\begin{equation}
\label{eqn_svr}
C \sum_{i=1}^N \mathcal{L}_{\epsilon}(y_i-\hat{y}_i) + \sum_{j=1}^M \theta_j^2
\end{equation}
where $\mathcal{L}_{\epsilon}$ is defined as
\begin{equation}
\mathcal{L}_{\epsilon}(y_i-\hat{y}_i)=\left\{\begin{array}{ll}
|y_i-\hat{y}_i|-\epsilon, & \text { if } |y_i-\hat{y}_i|>\epsilon \\
0, & \text { if } |y_i-\hat{y}_i|\leq\epsilon
\end{array}\right.
\end{equation}
Due to the relationship between the $C$ and $\epsilon$, $\epsilon$ is fixed while the other hyperparameters are tuned as suggested in \cite{kuhn2013applied}. 

The support vector methods can be applied to nonlinear problems, by projecting the input space into a Reproducing Kernel Hilbert Space, using a kernel function. In our analysis we apply the radial basis function (RBF) kernel (or squared exponential kernel), defined as:
\begin{equation}
\label{eqn_rbf}
\kappa(\bm{x}_i, \bm{x}_j) = \exp(-\gamma\|\bm{x}_i-\bm{x}_j\|^2)
\end{equation}
where $\bm{x}_i$ and $\bm{x}_j$ are input features in the original space and $\gamma$ is a kernel hyperparameter, tuned jointly with the cost parameter $C$.

\subsubsection{Multilayer Perceptron}
Multilayer perceptron (MLP) \cite{rosenblatt1961principles} is a class of feedforward artificial neural networks. The typical architecture consists of three types of layers: i) the input layer which received the input features $\bm{x}$, ii) an arbitrary number of hidden layers where the main computation is done and iii) the output layer which outputs the prediction. Each layer is formed by one or multiple operational units, e.g. neurons which compute a linear combination of the outputs from the previous layer. The nonlinearities are incorporated into the model by passing the output through an activation function. In our analysis, we applied the RELU activation function at the end of each hidden layer. Additionally, in the classification task, we applied the softmax \cite{goodfellow20166} activation function. To reduce the overfitting we employed the $L_2$ (weight decay) regularization. The training is done using the Adam optimizer \cite{kingma2014adam} with the adaptive learning rate, in batches of 200 samples. The training is stopped if the loss does not improve by $10^{-4}$ (tolerance) over 10 epochs. 

The architecture, e.g. the number of layers, the number of neurons together with the initial learning rate and the $L_2$ regularization strength are tuned in grid search.

%% file: chapters/results.tex
\section{Results}
\label{sec_results}
\begin{figure}[!htbp]
\centering
\subfloat[]{\includegraphics[width=0.45\linewidth, height = 4cm]{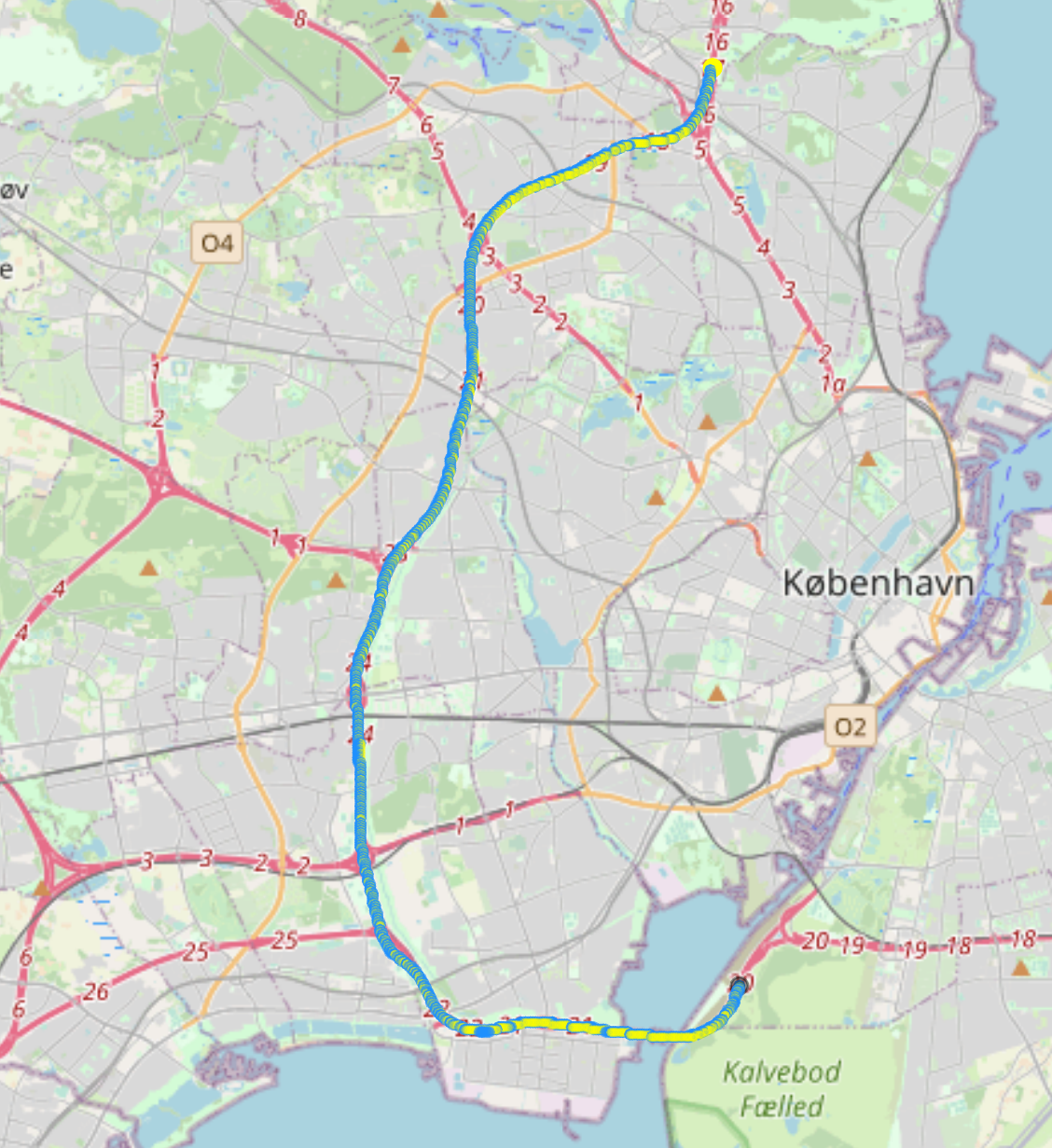}%
\label{fig_lin}}
\hfil
\subfloat[]{\includegraphics[width=0.45\linewidth, height = 4cm]{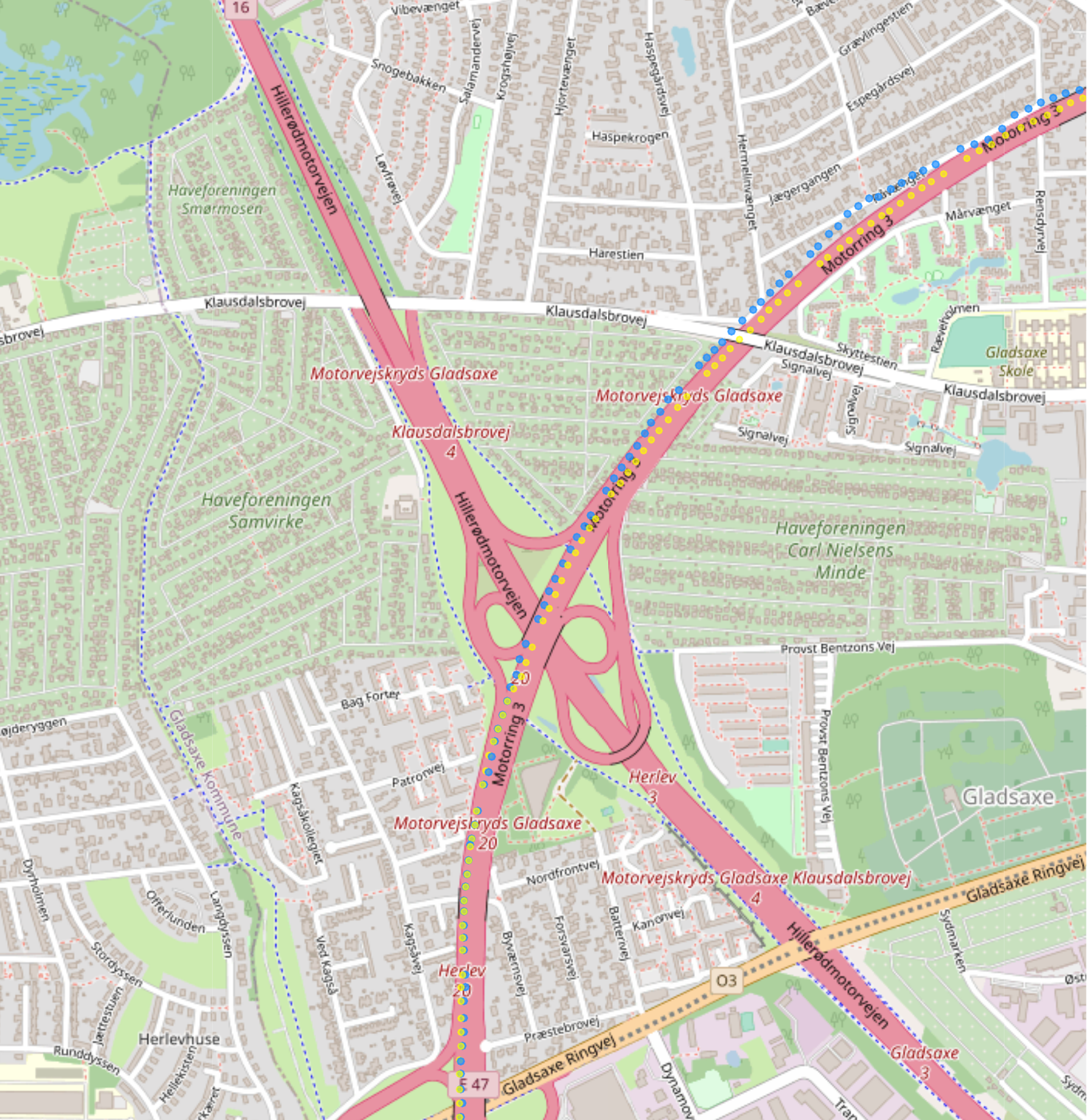}%
\label{fig_lasso}}
\caption{The full (left) route (in Copenhagen municipality) and the zoomed in (right) route before (blue) and after (yellow) map matching.}
\label{fig_matching}
\end{figure}

The analysis is performed on a \SI{50.3}{\km} long route, obtained by applying the segment alignment procedure and the sliding window approach. The obtained route and the impact of map matching are illustrated in Fig. \ref{fig_matching}. Due to the correlations between the segments, the random shuffled split can overestimate the performance - hence the final performance is evaluated on a holdout dataset which is defined as the last $20\%$ (1006 segments) section of the road. The first $80\%$ section (4025 segments) is exploited for training and hyperparameter tuning in the 5-fold cross-validation manner. 

\begin{figure}[!htbp]
\centering
\includegraphics[width=\linewidth]{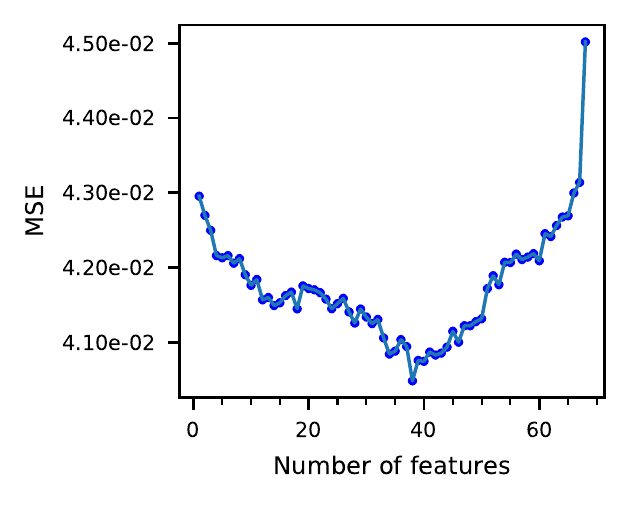}
\caption{The feature subset selection using the SFS.}
\label{fig:fs_sfs}        
\end{figure}

After the segments have been prepared, the $68$ features are computed in the temporal and statistical domain. Subsequently, the feature selection procedure is performed, including the removal of constant features and the SFS algorithm. The SFS feature performance is evaluated in a 5-fold cross-validation manner, selecting the feature giving the best mean performance across the folds. The SFS performance for different number for features selected  is shown in Fig. \ref{fig:fs_sfs}.

A subset of $38$ features is giving the best performance in terms of the $\mse$, given in Table \ref{table_feats} and detailed in \cite{barandas2020tsfel}. The machine learning models are tested on two different sets of features: i) the 38 features selected by SFS, and ii) the PCA projections of the SFS subset, selected to explain the \SI{99}{\%} of the variance, reducing the number of features to $23$.

\begin{table}[!htbp]
\renewcommand{\arraystretch}{1.1}
\caption{The Optimal Subset of Features Selected for Machine Learning.} 
\label{table_feats}
\centering
\begin{adjustbox}{width=0.9\linewidth}
\begin{tabular}{@{}lr@{}}
\toprule
                            \textbf{Extracted Feature} & \textbf{Sensor} \\
\midrule
                   Variance  & Acceleration-z  \\
            Mean of differences in the signal & Acceleration-z       \\
  Median of absolute deviations in the signal & Acceleration-z    \\
\SI{80}{\%} percentile value of the signal ECDF  & Acceleration-z    \\
 Median of differences in the signal   & Acceleration-z    \\
     Median & Acceleration-z    \\
  Mean of absolute deviations in the signal   & Acceleration-z    \\
 Median of absolute deviations in the signal & Vehicle speed \\
Median of differences in the signal & Vehicle speed \\
Mean & Acceleration-z    \\
 Number of positive turning points  & Acceleration-z    \\
The slope of the signal from a linear fit  & Acceleration-z    \\
 Standard deviation  & Acceleration-z    \\
 Number of negative turning points& Vehicle speed \\
Autocorrelation of the signal  & Acceleration-z    \\
Area under the curve of the signal  & Acceleration-z    \\
Root mean square of the signal  & Acceleration-z    \\
Centroid along the time axis  & Acceleration-z    \\
Absolute energy of the signal  & Acceleration-z    \\
Total energy of the signal  & Acceleration-z    \\
\SI{5}{\%} percentile value of the signal  & Acceleration-z    \\
 Median of absolute deviations in the signal  & Acceleration-z    \\
Kurtosis of the signal  & Acceleration-z    \\
Sum of absolute differences in the signal  & Acceleration-z    \\
Skewness of the signal  & Vehicle speed \\
Shannon Entropy of the signal  & Acceleration-z    \\
The total distance traveled by the signal  & Acceleration-z    \\
Number of negative turning points  & Acceleration-z    \\
         Number of neighbourhood peaks & Vehicle speed \\
\SI{80}{\%} percentile value of the signal ECDF   & Vehicle speed \\
        Interquartile range (\SI{75}{\%}-\SI{25}{\%}) & Acceleration-z    \\              
  Minimum  & Acceleration-z    \\
  Maximum  & Acceleration-z    \\
  Mean of absolute deviations in the signal  & Acceleration-z    \\
 \SI{20}{\%} percentile value of the signal ECDF & Acceleration-z    \\
       Peak to peak distance &               Vehicle speed \\
   Mean of absolute deviations in the signal & Vehicle speed \\
   Mean of differences in the signal  & Vehicle speed \\
\bottomrule
\end{tabular}
\end{adjustbox}
\end{table}


A set of regression and classification machine learning models is trained on the feature subset. The hyperparameter tuning is performed in a 5-fold cross validation. The folds are split so that the validation fold always follows the train folds, as appropriate to time series. 

\subsection{Regression}
In the regression task, the models are trained to predict the value of the IRI. As performance metrics, computed are the $R^{2}$, mean absolute error $MAE$, root mean square error $RMSE$ and mean relative error $MRE$. 

The multiple linear model has no tunable hyperparamaters, while in the regularized linear models, the regularization parameters are set at $0.05$ (Lasso) and $600$ (Ridge). In the elastic net, the L1 regularization parameter is set at $0.05$ and the l1 ratio is set at $0.2$, leading to a slight improvement than when using $L1$ regularization alone. In the kNN, the number of neighbors is set at 22, while in the random forest model, the number of trees is set at $400$, the maximum tree depth at $5$ and the maximum number of features at $6$. In the SVR model, the kernel coefficient $\gamma$ is set at $0.001$ and the cost parameter $C$ at $1$. In the MLP model, tested were multiple architectures with different sizes of layers and units. The best performance was obtained using the network with 3 hidden layers $(2, 4, 6)$, adaptive learning rate which starts at $0.01$ and L2 penalty set at $1$.

The performance is summarized in Table \ref{table:iri_reg_pred}. Since the baseline model is defined as a model which always predicts the mean value of the train dataset, due to a slight difference in the means between the train $\mu(IRI) = 1.23$ and the test dataset $\mu(IRI) = 1.22$, the baseline modelling results in a negative $R_2 = -0.15$. The linear model performs poorly when trained on the SFS subset due to its high sensitivity to the correlations between the input features, but a large improvement is obtained after reducing the multicollinearity with the PCA. In terms of all metrics, the best performance is obtained with the MLP model, both when trained using the SFS subset and the PCA transformed subset. The second best performance is obtained with the SVR, while other models show a slightly worse performance, but significantly better than the baseline model.

\begin{table}[!htbp]
\renewcommand{\arraystretch}{1.3}
\caption{The Performance Comparison of Machine Learning Models for IRI Prediction, Evaluated on Test Data.}
\label{table:iri_reg_pred}
\centering
\begin{adjustbox}{width=\linewidth}
\begin{tabular}{@{}lllllllll@{}}
\toprule
 \multirow{2}{*}{\textbf{Model}} & \multicolumn{4}{c}{\textbf{Without Dimension Reduction}} & \multicolumn{4}{c}{\textbf{PCA}}\\
    \cline{2-5}  \cline{6-9}
     & $\textbf{R}^\textbf{2}$ & \textbf{MAE} & \textbf{RMSE} & \textbf{MRE} & $\textbf{R}^\textbf{2}$ & \textbf{MAE} & \textbf{RMSE} & \textbf{MRE}\\
\midrule
           Baseline &      -0.15 &          0.33 &            0.38 &           0.36 &         -0.15 &           0.33 &            0.38 &           0.36 \\
 Linear &         -0.64 &           0.37 &            0.46 &           0.39 &          0.57 &           0.19 &            0.24 &           0.20 \\
           Lasso &          0.57 &           0.19 &            0.23 &           0.21 &          0.54 &           0.19 &            0.24 &           0.21 \\
           Ridge &          0.57 &           0.19 &            0.23 &           0.20 &          0.57 &           0.19 &            0.24 &           0.20 \\
            Elastic Net &           0.57 &         0.19 &          0.23 &         0.20 &              0.56 &               0.19 &                0.24 &               0.20 \\
K-Nearest Neighbors &        0.59 &           0.18 &            0.23 &           0.20 &          0.59 &          0.19 &           0.23 &           0.20 \\
   Random Forest &          0.58 &           0.19 &            0.23 &           0.20 &          0.53 &           0.20 &            0.25 &           0.22 \\
             SVR &          0.63 &           0.18 &            0.22 &           0.19 &          0.62 &           0.18 &            0.22 &           0.19 \\
             MLP &  \textbf{0.64} & \textbf{0.17} &   \textbf{0.21} &    \textbf{0.18} &\textbf{0.66} & \textbf{0.16} &   \textbf{0.21} &   \textbf{0.16} \\
\bottomrule
\end{tabular}
\end{adjustbox}
\end{table}

\begin{figure*}[!t]
\centering
\subfloat[Linear]{\includegraphics[width=0.23\linewidth]{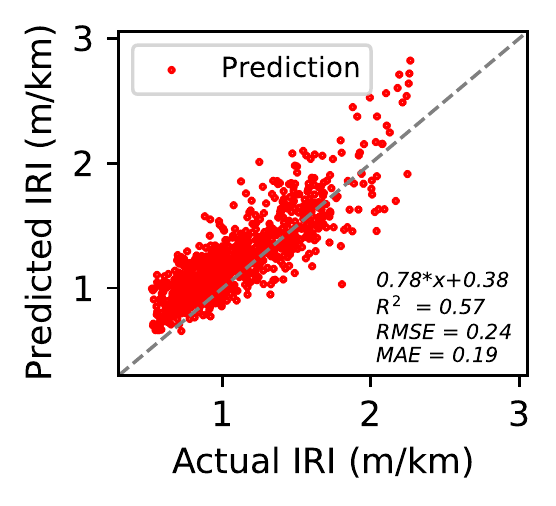}}
\hfil
\subfloat[Lasso]{\includegraphics[width=0.23\linewidth]{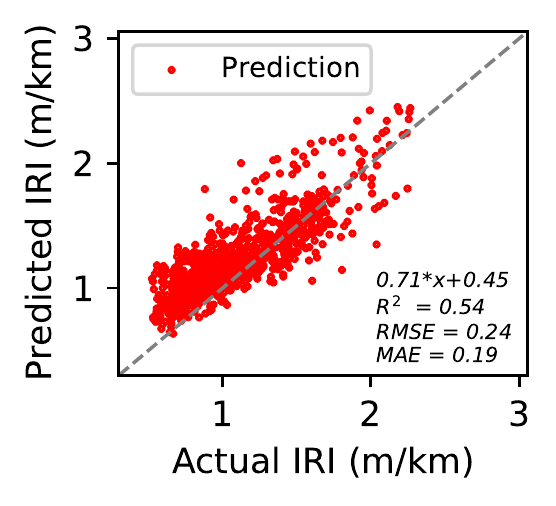}}
\hfil
\subfloat[Ridge]{\includegraphics[width=0.23\linewidth]{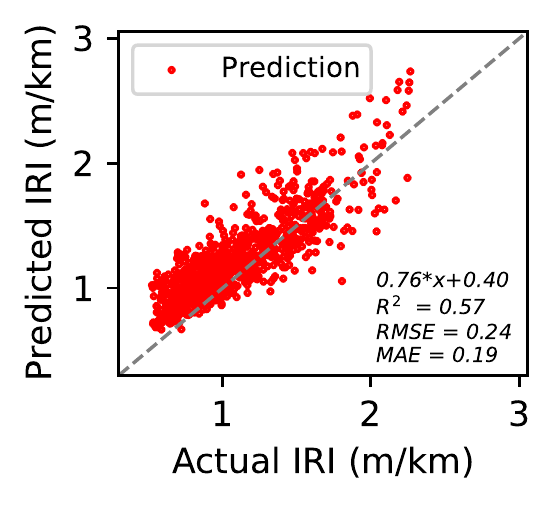}}
\hfil
\subfloat[Elastic Net]{\includegraphics[width=0.23\linewidth]{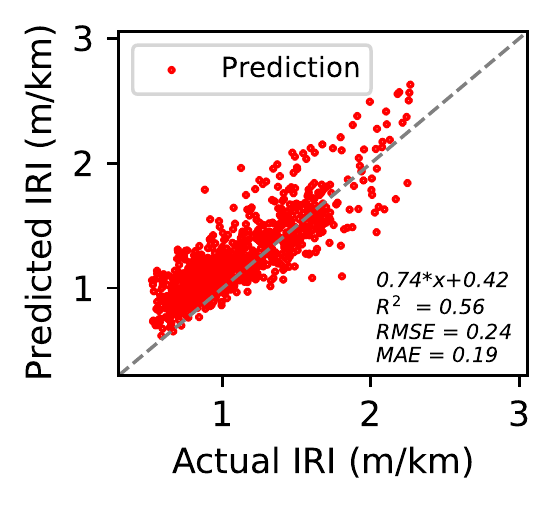}}\\
\subfloat[k-Nearest Neighbors]{\includegraphics[width=0.23\linewidth]{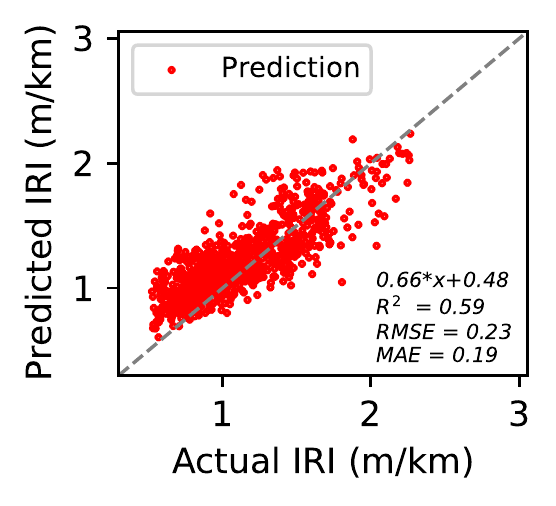}}
\hfil
\subfloat[Random Forest]{\includegraphics[width=0.23\linewidth]{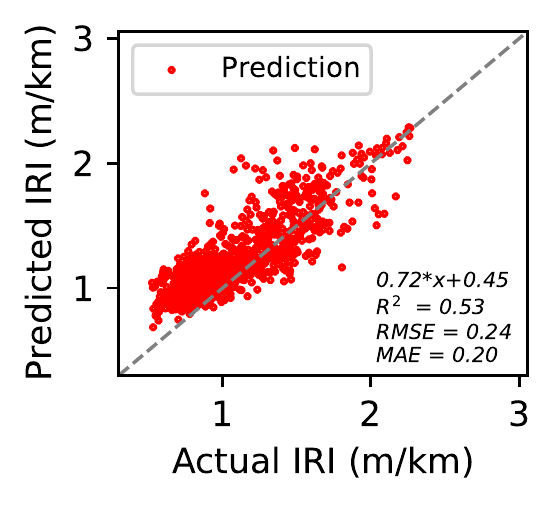}}
\hfil
\subfloat[SVR]{\includegraphics[width=0.23\linewidth]{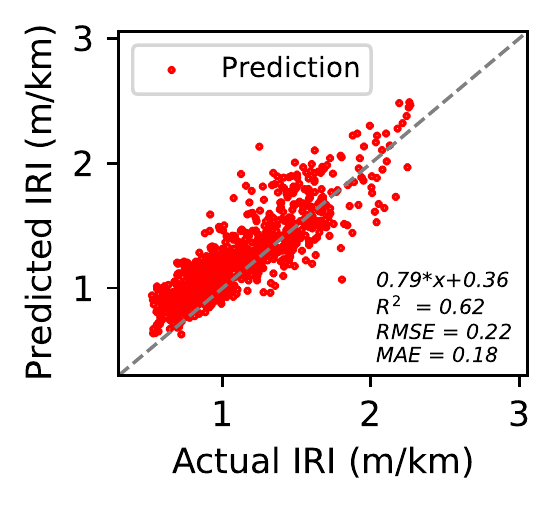}}
\hfil
\subfloat[MLP]{\includegraphics[width=0.23\linewidth]{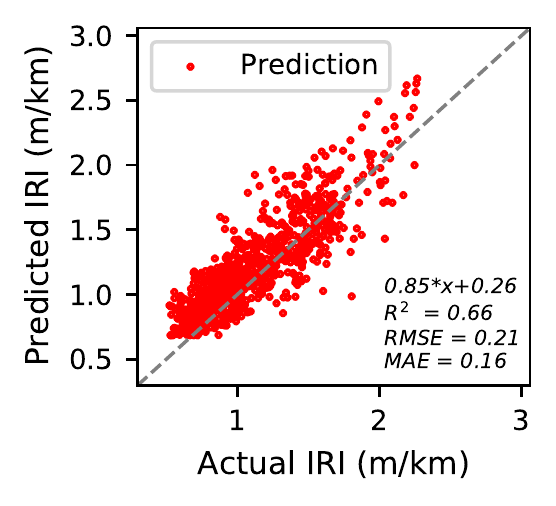}}
\caption{The performance comparison of ML models for IRI prediction. Shown are the actual versus predicted values. The models are trained on the PCA components and evaluated on test data.}
\label{fig_irireg}
\end{figure*}

The predicted versus the actual IRI values for models, trained on the PCA transformed subset are presented in Fig. \ref{fig_irireg}. All models show a good prediction trend, since majority of predictions are near the diagonal. Observed is a slightly worse performance for higher values of IRI, which is expected due to the smaller number of samples with high IRI.

\subsection{Classification}


In the classification task, a set of ML models is trained to predict the IRI severity. As performance metrics we compute the $Precision$, $Recall$ and $F_{1}\mathdash score$ and averaged over all classes. The performance is summarized in Table \ref{table:iri_class_pred}, with the best results per each metric being showed in bold face. The best performance in terms of the $F_{1}\mathdash score$ is obtained with the logistic regression model when trained on the SFS subset. When trained on the PCA transformed subset, the best performance is obtained with the SVM model followed by the logistic regression and the MLP. The highest $Recall=0.74$, with $Precision=0.65$, is obtained using the logistic regression and the SVM models. All models result in a significantly better performance in terms of all metrics than the baseline model. 

\begin{figure*}[!htbp]
\centering
\subfloat[Naive Bayes]{\includegraphics[width=0.3\linewidth]{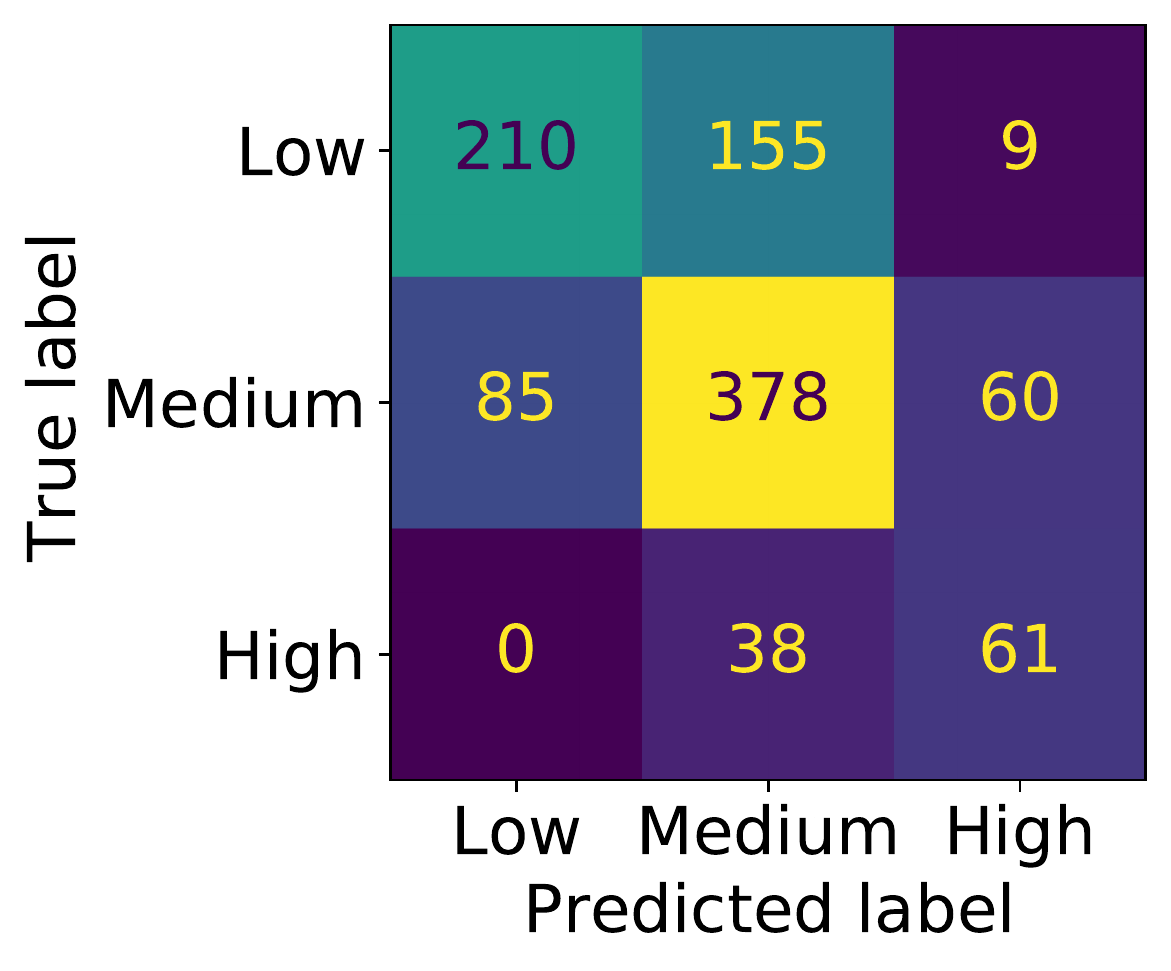}}
\hfil
\subfloat[K-Nearest Neighbors]{\includegraphics[width=0.3\linewidth]{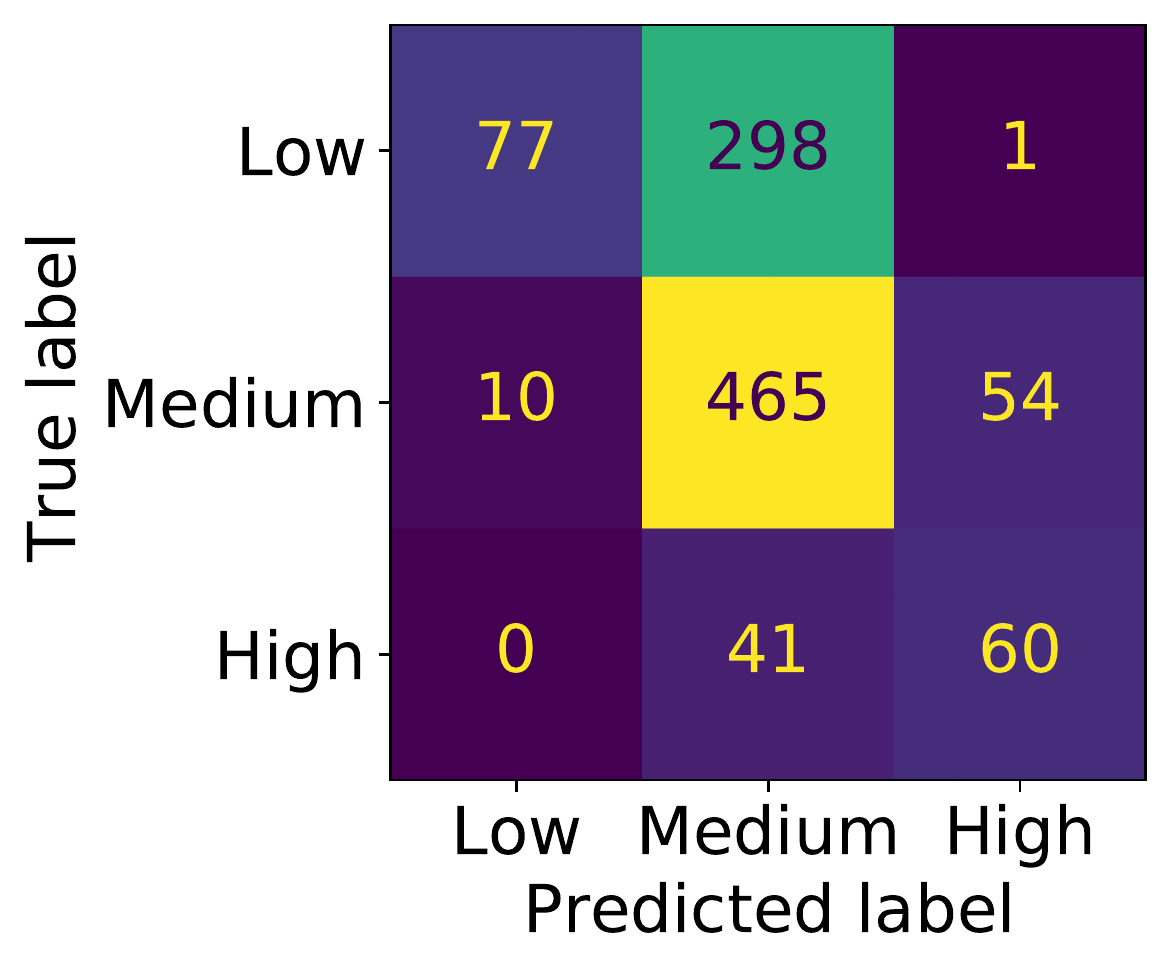}}
\hfil
\subfloat[Logistic Regression]{\includegraphics[width=0.3\linewidth]{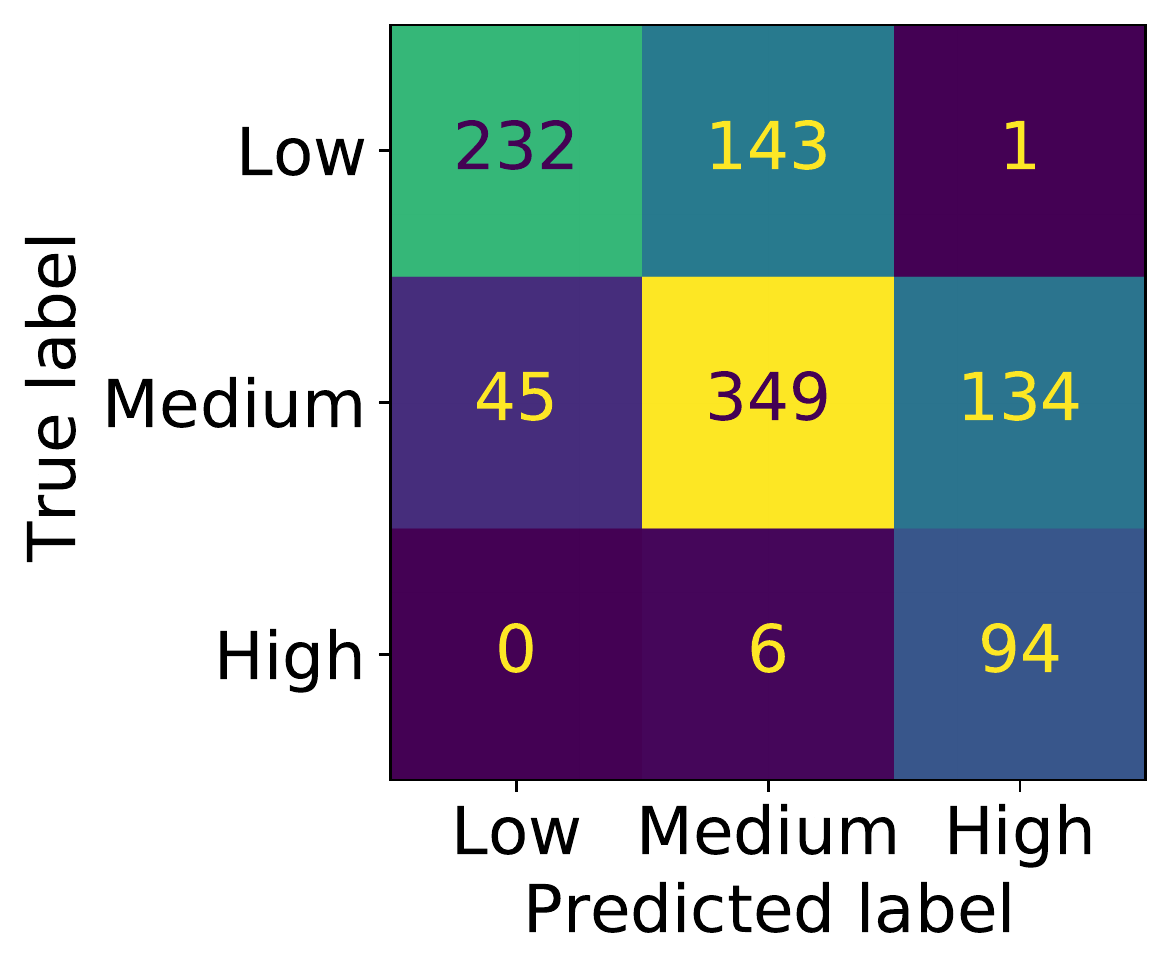}}\\

\subfloat[Random Forest]{\includegraphics[width=0.3\linewidth]{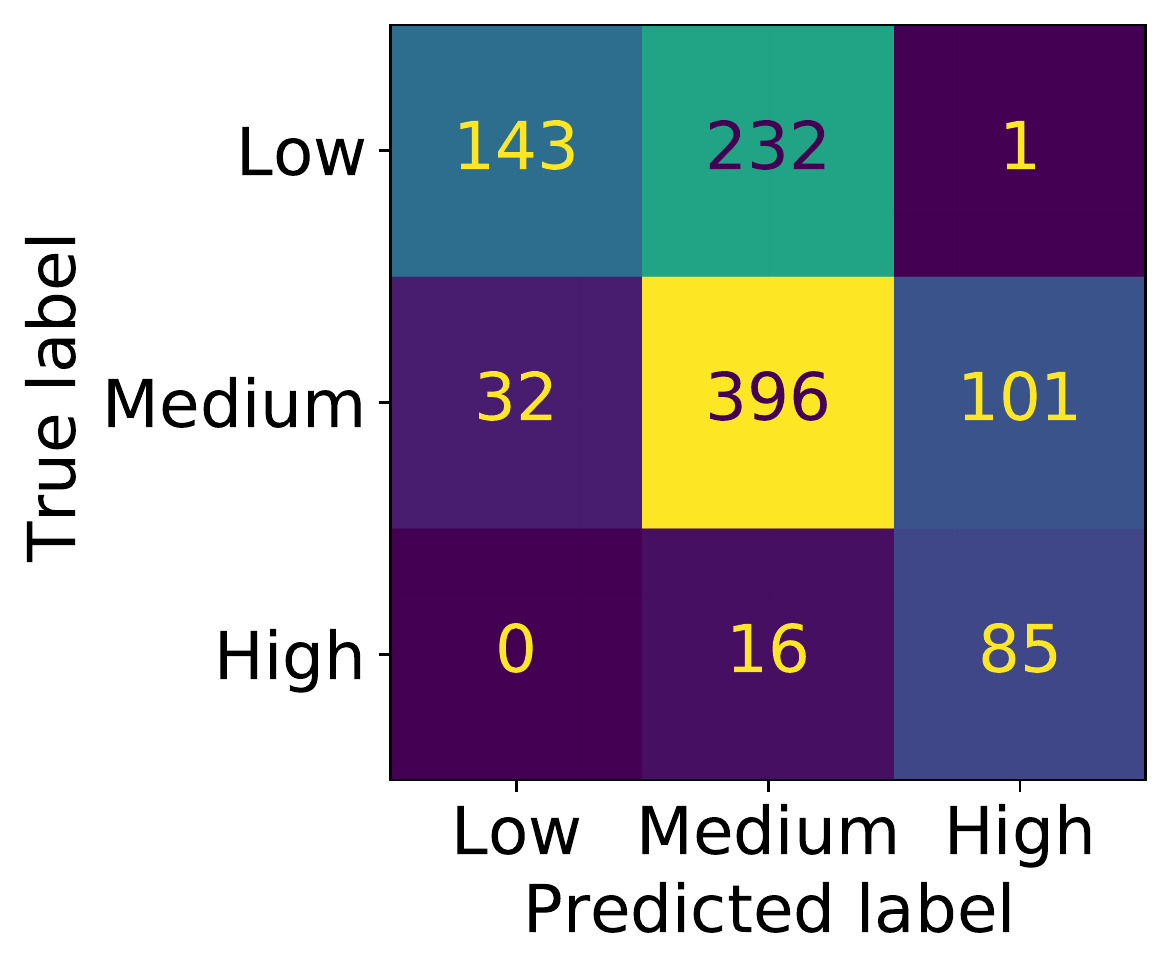}}
\hfil
\subfloat[SVC]{\includegraphics[width=0.3\linewidth]{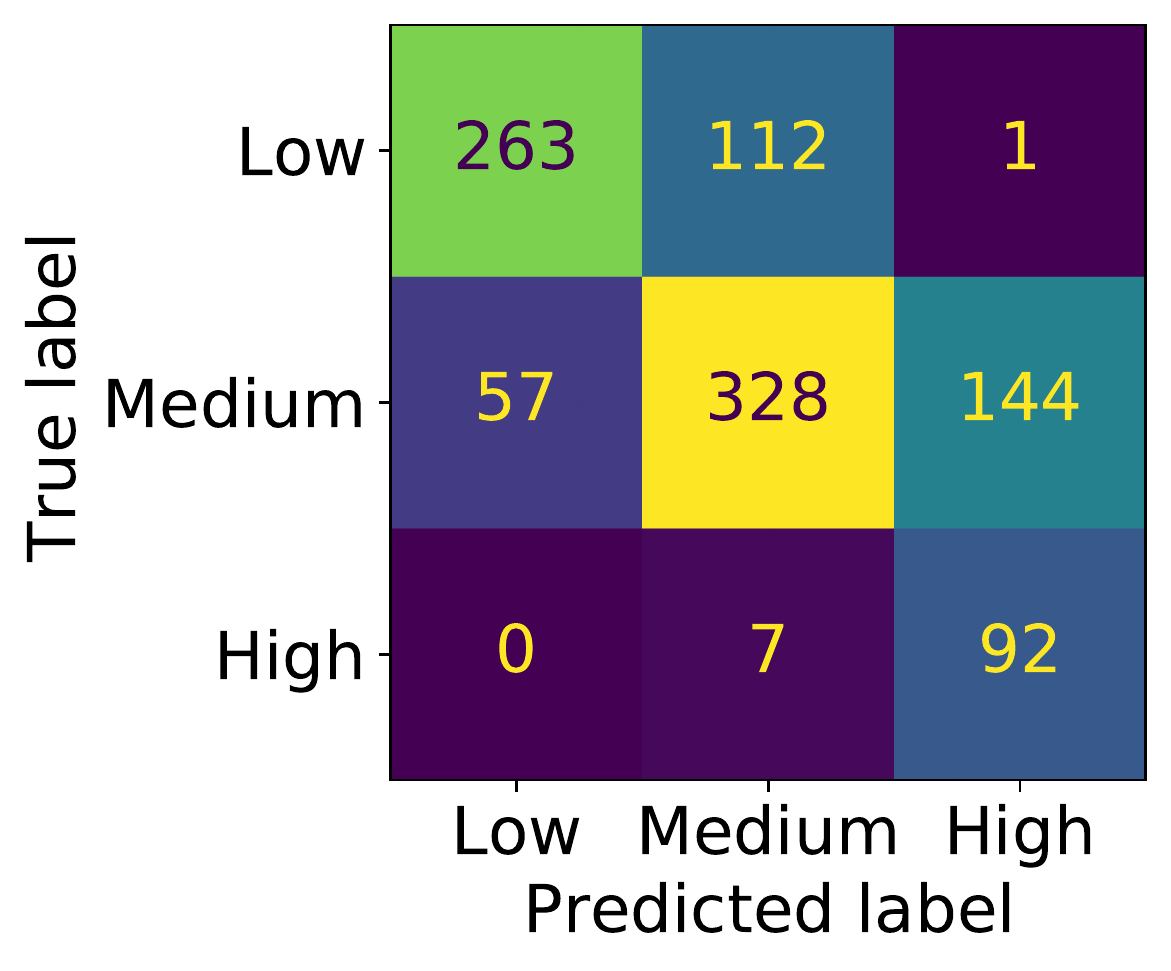}}
\hfil
\subfloat[MLP]{\includegraphics[width=0.3\linewidth]{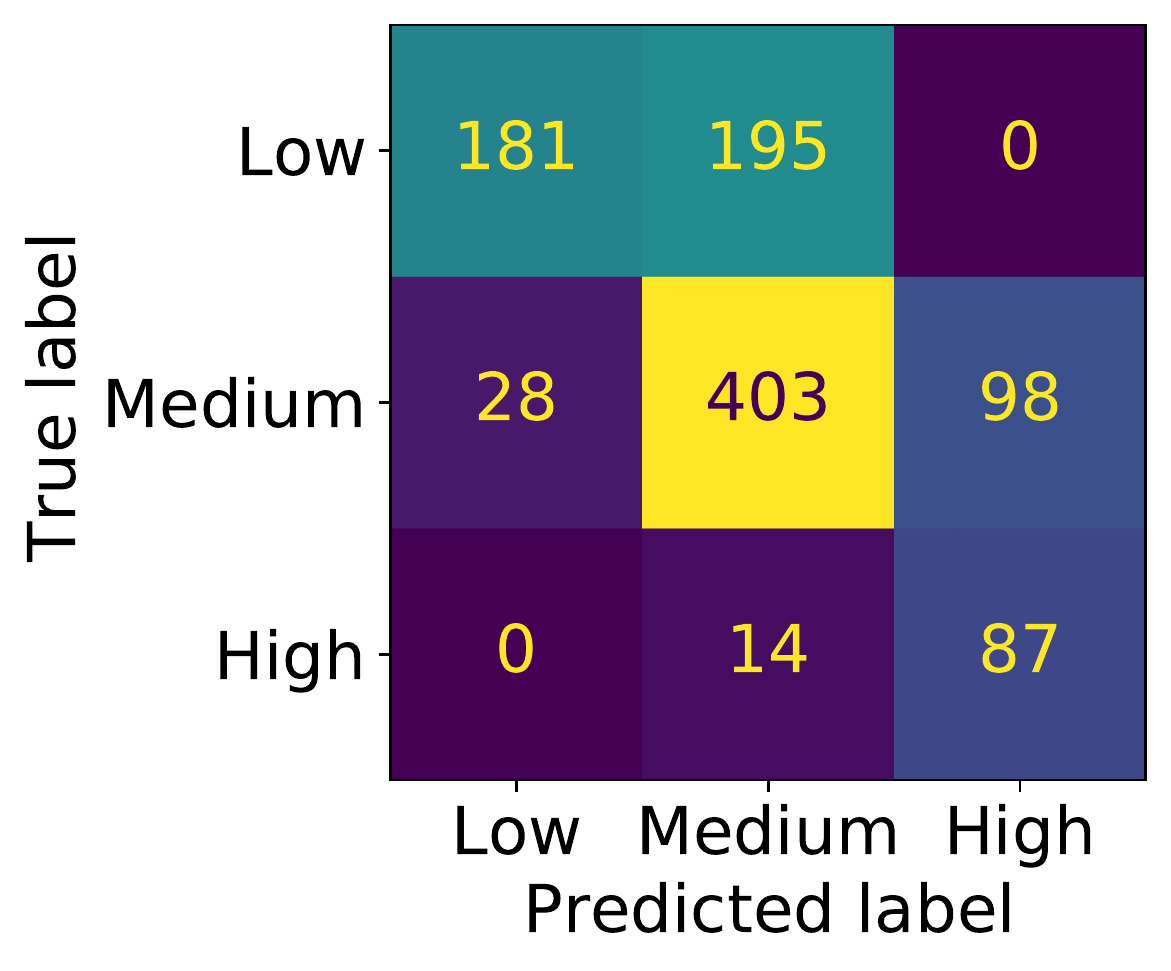}}\\
\caption{The performance comparison of ML models for IRI class prediction, visualized per each class. Shown are the true versus predicted labels. The models are trained on the PCA components and evaluated on test data.}
\label{fig_iriclass}        
\end{figure*}

The performance evaluation for different classes is visualized via confusion matrices, shown in Fig. \ref{fig_iriclass} for the models trained on the PCA transformed subset. The rows show the true class, while the columns show the predicted class. A good performance of our models is indicated by the fact that the majority of all predictions fall onto the diagonal, indicating the correct predictions and that the mislabeled classes are in majority of cases assigned to the neighboring class (the classes are on an ordinal scale).

\begin{table}[!htbp]
\renewcommand{\arraystretch}{1.3}
\caption{The Performance Comparison of Machine Learning Models for IRI Level Prediction, Averaged over all Classes and Evaluated on Test Data.}
\label{table:iri_class_pred}
\centering
\begin{adjustbox}{width=\linewidth}
\begin{tabular}{@{}lllllll@{}}
\toprule
 \multirow{2}{*}{\textbf{Model}} & \multicolumn{3}{c}{\textbf{Without Dimension Reduction}} & \multicolumn{3}{c}{\textbf{PCA}}\\
    \cline{2-4}  \cline{5-7}
      &\textbf{Precision} & \textbf{Recall} & \textbf{F1-Score} & \textbf{Precision} & \textbf{Recall} & \textbf{F1-Score} \\
\midrule
              Baseline &            0.18 &            0.33 &              0.23 &                     0.18 &                  0.33 &                    0.23 \\
              Logistic Regression &               0.67 &   \textbf{0.76} &     \textbf{0.67} &                     0.65 &          \textbf{0.74}&                    0.65 \\
                K-Nearest Neighbors &               0.67 &            0.56 &              0.53 &            \textbf{0.66} &                  0.56 &                    0.53 \\
        Naive Bayes &               0.48 &            0.52 &              0.50 &                     0.61 &                  0.63 &                    0.61 \\
     Random Forest &               0.65 &            0.66 &              0.59 &                     0.63 &                  0.66 &                    0.59 \\
                SVM &               0.65 &            0.75 &              0.66 &                     0.65 &          \textbf{0.74}&            \textbf{0.66} \\
                MLP &               \textbf{0.67} &   0.59 &              0.51 &            \textbf{0.66} &                  0.70 &                    0.64 \\
\bottomrule
\end{tabular}
\end{adjustbox}
\end{table}

%% file: chapters/conclusion.tex
\section{Conclusion}
\label{sec_conc}

We have presented a machine learning pipeline for road roughness prediction, developed using the car acceleration in the $z$-direction and a varying car speed, recorded by the in-vehicle sensor. The dataset included \SI{50.3}{\km} long road section along the M3 Motorway in Copenhagen. We developed a preprocessing pipeline, including the map matching of the vehicle trajectory to the OSM road network based on the HMM model, and implemented using the OSRM service and segment alignment using the GPS sensor data. The predicting features are computed in the temporal and statistical domain from the sensor data and an optimal subset is selected using the SFS. Traditional machine learning methods were compared including the multivariate ordinary and regularized linear models, naive Bayes, k-nearest neighbors, random forest, support vector machines, and the multi-layer perceptron neural network. The models were trained both to predict the exact value and the severity of the IRI, averaged over \SI{100}{\m}. The models were tuned in cross-validation manner and evaluated on an independent holdout test dataset, resulting in a performance of $R^2=0.66$ (MLP) and $\rmse=0.21$ (MLP) for IRI value prediction, and for classification we achieved a $recall=0.74$ with $precision=0.65$ (logistic regression, SVM) and $\f=0.66$ (SVM). The results demonstrate that the proposed pipeline can be used for accurate IRI prediction, hence enabling timely monitoring of road network, using affordable in-vehicle sensors.